\documentclass{article}
\usepackage{spconf,graphicx,hyperref}
\usepackage{multirow}
\usepackage{makecell}
\usepackage{array}
\usepackage{tabularx}
\usepackage{booktabs}
\usepackage{amssymb}
\usepackage{amsthm,amsmath,pifont}
\usepackage{mathrsfs}
\usepackage[table]{xcolor}
\usepackage{comment}
\usepackage{caption}

\title{Modality-Decoupled RGB-Thermal Object Detector via Query Fusion}
\name{Chao Tian, Zikun Zhou*, Chao Yang, Guoqing Zhu, Fu'an Zhong, Zhenyu He* \thanks{* Corresponding Author. Email: zhouzikunhit@gmail.com, zhenyuhe@hit.edu.cn}}
\address{Harbin Institute of Technology, Shenzhen}
\begin{document}
\def\eg{\emph{e.g.},}
\def\etal{\emph{et al.}}
\def\ie{\emph{i.e.},}
\def\etc{\emph{etc.}}
\def\wrt{w.r.t.}
\def\vs{\emph{vs.}}
\def\blue#1{\textcolor{blue}{#1}}
\def\red#1{\textcolor{red}{#1}}
\def\green#1{\textcolor{green}{#1}}

\makeatletter
\def\va@prefix{va:}

\DeclareRobustCommand{\setword}[1]{%
  \phantomsection
  \edef\@currentlabel{\unexpanded{#1}}
  \label{\va@prefix#1}#1%
}

\DeclareRobustCommand{\varef}[1]{%
  \hyperref[\va@prefix#1]{\ref*{\va@prefix#1}}%
}
\makeatother

\ninept
\maketitle
\begin{abstract}
The advantage of RGB-Thermal (RGB-T) detection lies in its ability to perform modality fusion and integrate cross-modality complementary information, enabling robust detection under diverse illumination and weather conditions. However, under extreme conditions where one modality exhibits poor quality and disturbs detection, modality separation is necessary to mitigate the impact of noise. To address this problem, we propose a Modality-Decoupled RGB-T detection framework with Query Fusion~(MDQF) to balance modality complementation and separation. In this framework, DETR-like detectors are employed as separate branches for the RGB and TIR images, with query fusion interspersed between the two branches in each refinement stage. Herein, query fusion is performed by feeding the high-quality queries from one branch to the other one after query selection and adaptation. This design effectively excludes the degraded modality and corrects the predictions using high-quality queries. Moreover, the decoupled framework allows us to optimize each individual branch with unpaired RGB or TIR images, eliminating the need for paired RGB-T data. Extensive experiments demonstrate that our approach delivers superior performance to existing RGB-T detectors and achieves better modality independence.
\end{abstract}
\begin{keywords}
RGB-T, Object Detection, Thermal Infrared, Multispectral, Query Fusion
\end{keywords}

\vspace{-1mm}
\section{Introduction}
\label{sec:intro}

Traditional visible-spectrum images~(\ie~RGB images) often lack robustness under extreme conditions, such as low-light environments~\cite{ku2018joint} and adverse weather conditions~(\eg~including fog and rain)~\cite{zhang2020multispectral_flir}. These factors degrade image quality and consequently undermine the robustness and accuracy of object detection. To deal with this problem, many studies introduce thermal infrared images for complementary information, which capture thermal radiation and are resilient to lighting conditions~\cite{liu2016multispectral}. By leveraging both RGB and thermal infrared images, RGB-Thermal (RGB-T) object detection shows its robustness under challenging environmental conditions and receives increasing attention in the object detection community~\cite{zhang2020multispectral_flir, xiong2025efficient, m3fd} for improving both accuracy and robustness.

The advantage of RGB-T object detection stems from the complementary characteristics of the two modalities~\cite{yuan2024improving, tian2023cross, zhang2019weakly}. Consequently, efficient cross-modality fusion is essential for RGB-T detection, although it remains inherently complex. In typical scenarios where both RGB and TIR images possess satisfactory quality, the objective of fusion is to exploit the complementary information from both modalities, thereby generating more discriminative features to distinguish the objects. Conversely, under extreme conditions, such as low-contrast illumination, the contributions of the RGB and TIR modalities to object detection can become highly imbalanced, as the degraded modality would introduce noise during fusion~\cite{mbnet}. This issue is also known as \textit{modality imbalance}. When such a modality imbalance arises solely during testing, the detector encounters an out-of-distribution (OOD) challenge~\cite{Tian_Yang_Zhu_Wang_He_2025}, which may result in performance inferior to that of single-modality detectors. In summary, modality complementation is crucial in regular conditions, whereas modality separation should be prioritized in extreme cases. Thus, it is imperative for RGB-T detectors to effectively balance modality complementation and separation during fusion.
Existing RGB-T detection methods either tightly couple the two modalities~\cite{Zhang_TIV, mbnet, zhang2019weakly, kim2021uncertainty}, making them sensitive to modality imbalance, or maintain a high level of modality separation at the cost of losing cross-modality complementarity~\cite{chen2022multimodal}.

{
\setlength{\textfloatsep}{10pt}
\begin{figure}[t]
    \centering
    \includegraphics[width=\linewidth]{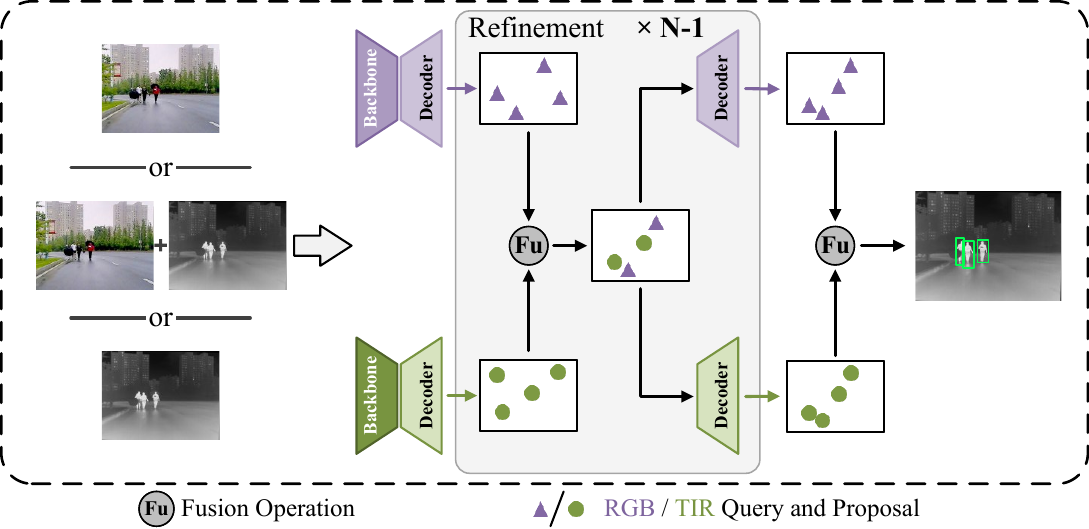}
    \captionsetup{skip=3pt}
    \caption{Our proposed RGB-T detector equipped with the query fusion strategy for cross-modality information exchange. The framework keeps the independence of each branch, avoiding failure during the degradation of a modality. And each single-modality branch can be optimized separately.}   
    \label{fig:moti}
\end{figure}
}

{
\setlength{\textfloatsep}{10pt}
\begin{figure*}[t]
\centering
\includegraphics[width=\linewidth]{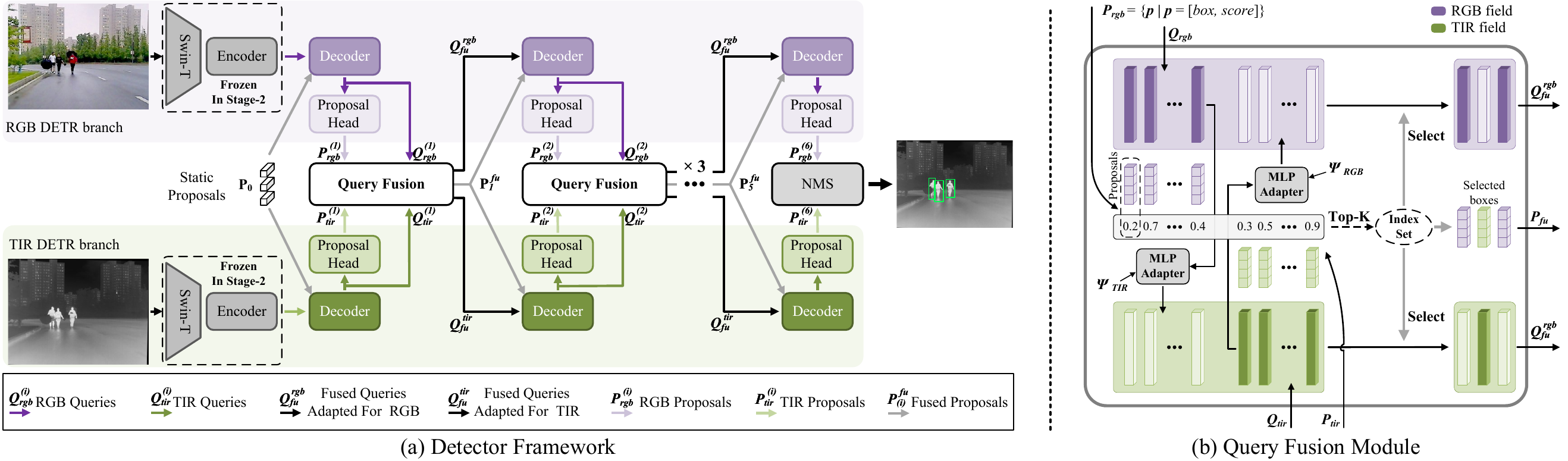}
\captionsetup{skip=3pt}
\caption{(a) MDQF framework. Two standard DETR-like detectors are deployed to process RGB and TIR images independently. The continuous optimization of proposals in decoders of each modality is influenced by the other modality through the query fusion. (b) Query Fusion module. This is a practical implementation of the query fusion, including projection, selection, and collection of query vectors. 
}
\label{fig:method}
\end{figure*}
}

To effectively balance modality complementation and separation, we propose a Modality-Decoupled detection framework with Query Fusion~(MDQF), as shown in Figure~\ref{fig:moti}. MDQF is designed to adhere to two principles: 1)~effective cross-modality information prompting; and 2)~preservation of the original architecture and parameters of each single-modality branch. In this framework, DETR-like detectors~\cite{detr} are employed as separate branches for the RGB and TIR modalities, with query fusion interspersed between the two branches in each refinement stage.
Specifically, the query fusion mechanism operates after each decoder layer to fuse queries and proposals from DETR-like detectors. First, it chooses the top-\textit{k} queries and the corresponding proposals based on proposal confidence using a customized selection module. Then it employs a lightweight query adapter to align the high-quality queries from one branch to the decoder of the other branch. With the adapted high-quality queries, each branch independently predicts enhanced detection results, which are further combined. Thus, query fusion achieves the cross-modality complementation while selectively filtering out low-quality queries/proposals to prevent them from affecting the following decoders. This design effectively excludes the degraded modality and corrects predictions in the corresponding branch using high-quality prompts. By preserving the original branches, the framework remains functional with the individual branch, even when one modality is entirely unavailable. 

We employ a separate-to-joint learning strategy to preserve branch independence while effectively fusing modalities. Benefiting from the modality independence brought by the query fusion and the training strategy, each branch of the MDQF framework can be optimized separately, which alleviates the requirement of paired RGB-T image data. 
Our contributions are summarized as follows:
\begin{itemize}
\item We propose a query fusion strategy to effectively balance the modality complementation and modality separation during cross-modality fusion.  
\item  We propose a modality-decoupled RGB-T detection framework based on the query fusion, in which the two branches can perform detection collaboratively, while each branch can also conduct detection or optimization independently. 
\item Extensive experimental results demonstrate the superior performance of our approach compared with state-of-the-art RGB-T detectors, and its robustness to modality degradation. 
\end{itemize}

\vspace{-1mm}
\section{Method}
\vspace{-1mm}
\subsection{Overview of the Detection Framework} \vspace{-1mm}
Figure~\ref{fig:method} (a) illustrates the overall architecture of our modality-decoupled detection framework. It consists of two separate DETR-based detectors, each dedicated to a single modality. The prediction of the $i$-th decoder layer in DETR is formulated as:
\begin{alignat}{1}
    \mathbf{Q}_i &= Decoder_i(\mathbf{v}, \mathbf{Q}_{i-1}, \mathbf{P}_{i-1}), \label{eq:decoder} \\ 
    \mathbf{P}_{i} &= Head_i(\mathbf{Q}_i, \mathbf{P}_{i-1}), i \in [1,6], \label{eq:head}
\end{alignat}
where $\textbf{v}$, $\textbf{Q}$, and $\textbf{P}$ respectively mean the value, query, and proposal of DETR. 
The core idea is to enable complementary information from one modality to enhance the other by refining the sets of intermediate proposals $\textbf{P}$ and queries $\textbf{Q}$ within each DETR branch. This exchange does not alter the original parameters or architectures of the individual branches, thereby preserving their independence. We refer to this process as query fusion.

As shown in Figure~\ref{fig:method} (a), at each decoder stage, the sets of proposals and queries from both branches are fused and refined for the next stage by query fusion $\mathscr{F}(\cdot)$, formulated as:
\begin{equation} \label{eq:qry_fu}
    \mathbf{P}^{(i-1)}_{fu}, \mathbf{Q}^{(i-1)}_{fu} = \mathscr{F}(\mathbf{P}^{(i-1)}_{rgb}, \mathbf{P}^{(i-1)}_{tir}, \mathbf{Q}^{(i-1)}_{rgb}, \mathbf{Q}^{(i-1)}_{tir}).
\end{equation}
Then, guided by the fused set of proposals $\mathbf{P}^{(i-1)}_{fu}$ and queries $\mathbf{Q}^{(i-1)}_{fu}$, each DETR branch independently conducts a new round of object prediction according to Eq.~\eqref{eq:decoder} and Eq.~\eqref{eq:head}, to enhance both the reliability and precision of the detection. Due to the separate predictions in each modality, the detection results require post-processing using non-maximum suppression~(NMS).

Our proposed framework maintains the independence of each single-modality branch. As a result, it can operate as a standalone DETR-like detector when one modality is missing during testing, thereby mitigating failures caused by out-of-distribution (OOD) inputs. Moreover, each branch can be optimized independently using single-modality data, which is practically beneficial as it reduces the reliance on paired RGB-T training data. 

\vspace{-1mm}
\subsection{Query Fusion} \vspace{-1mm}
We introduce the query fusion to conduct the fusion operation $\mathscr{F}(\cdot)$ in Eq.~\eqref{eq:qry_fu}. There are two major steps to conduct query fusion.

\noindent
\textbf{Query Selection.} 
The query fusion procedure collects high-quality and complementary queries and proposals from different modalities while removing those with low confidence. In DETR-like detectors, a query is explicitly associated with a proposal, which is predicted based on this query and indicates the location and class information of the query. We thus adopt the top-k strategy to select queries according to the confidence of their associated proposals. 

Each proposal $p=[cx,cy,w,h,s]$ is a vector that contains a bounding box and a score. Then we select the top-\textit{k} best proposals and queries among those from both RGB and TIR modalities according to their score $s$:
The procedure is formulated as:
\begin{alignat}{1}
    \mathbf{P}_{fu}, Z &= \mathop{TopK}([\mathbf{P}_{rgb}, \mathbf{P}_{tir}], k), \label{eq:topk} \\
    \mathbf{Q}_{fu} &= [\mathbf{Q}_{rgb}, \mathbf{Q}_{tir}](Z). \label{eq:select}
\end{alignat}
where $[\mathbf{P}_{rgb}, \mathbf{P}_{tir}]$ means the union set of proposals from both modalities, $Z \in \mathbb{N}^{1 \times k}$ is the index of the \textit{k} best proposals.

\noindent
\textbf{Query Adaptation.}
The $\mathbf{Q}_{fu}$ in Eq.~\eqref{eq:select} contains queries from two DETR branches, which are different in modes and distributions. The subsequent decoder layer~(Eq.~\eqref{eq:decoder}) needs tuning to handle the imported queries.
This procedure alters the original parameters of each vanilla DETR detector, thereby compromising the independence of individual branches. We deploy several additional lightweight modules, \ie~the adapter, alongside two DETR branches to project the queries, fitting them from one modality to the other. Each decoder stage is equipped with a pair of unshared adapters, each for a branch, performing the directional projection independently, as shown in Figure~\ref{fig:method} (b). 
A lightweight multilayer perceptron~(MLP) is employed to process the projection, denoted as $\Psi(\cdot)$. 

In our fusion design, the number of queries selected per modality is dynamic, which hinders efficient computation on parallel hardware due to variable tensor shapes. To address this, we perform query projection before selection, ensuring that both input and output tensors maintain static dimensions. Eq.~\eqref{eq:select} is thus modified as:
\begin{equation} \label{eq:select_rgbt}
\begin{alignedat}{3}
    &\mathbf{Q}^{rgb}_{fu}& &= [\mathbf{Q}_{rgb},~\Psi_{RGB}(\mathbf{Q}_{tir})& &](Z), \\
    &\mathbf{Q}^{tir}_{fu}& &= [\Psi_{TIR}(\mathbf{Q}_{rgb}),~\mathbf{Q}_{tir}& &](Z).
\end{alignedat}
\end{equation}
This design preserves computational efficiency and facilitates hardware acceleration. The $\mathbf{Q}_{fu}$ in Eq.~\eqref{eq:decoder} needs a replacement by $\mathbf{Q}^{rgb}_{fu}$ and $\mathbf{Q}^{tir}_{fu}$ respectively for different decoders, while the $\mathbf{P}_{fu}$ is shared for both modalities. The procedure of Eq.~(\ref{eq:decoder} -- \ref{eq:head}) is formulated as:
\begin{alignat}{1}
    \mathbf{Q}^{(i)}_{m} &= Decoder^{m}_{(i)}(\mathbf{v
    }_{m},[\mathbf{Q}^{m}_{fu}]^{(i-1)},\mathbf{P}^{(i-1)}_{fu}), \\
    \mathbf{P}^{(i)}_{m} &= Head^{m}_{(i)}(\mathbf{Q}^{(i)}_{m}, \mathbf{P}^{(i-1)}_{fu}),  m \in \{rgb, tir\}.
\end{alignat}

\vspace{-1mm}
\subsection{Separate-to-Joint Model Learning} \vspace{-1mm}

\textbf{Separate-to-Joint Training.}
Maintaining the independence of each single-modality branch is a fundamental requirement of our framework, while cross-modality complementarity is also essential. We thus adopt a two-stage training strategy. As illustrated in Figure~\ref{fig:method} (a), two DETR detectors are trained separately in the first stage, each for a modality. The training procedure follows the standard DETR paradigm.
Parameters of the two branches are used to initialize the second stage as a checkpoint after convergence. In this stage, the full framework is jointly trained on paired and aligned RGB-T images for a few epochs, with all backbone and encoder layers frozen. In subsequent optimization, the entire process is repeated in a loop of separate training, parameter update, and joint training.

\noindent
\textbf{Total Loss.}
The total loss of each single-modality DETR detector with 6 decoder layers can be written as:
\begin{equation}
    \mathcal{L} = \sum^{6}_{i=1} \alpha \mathcal{L}^{(i)}_{cls} + \beta \mathcal{L}^{(i)}_{iou} + \gamma \mathcal{L}^{(i)}_{L1},
\end{equation}
The $\mathcal{L}_{cls}$ is the classification loss, which is the Binary Cross-Entropy~(BCE) Loss. $\mathcal{L}_{iou}$ and $\mathcal{L}_{L1}$ are constraints of box regression, where $\mathcal{L}_{iou}$ is the GIoU loss. The hyper-parameters $\alpha$, $\beta$, and $\gamma$ are respectively set to 0.125, 0.25, and 0.625. The total loss of the whole framework in fine-tuning is the sum of losses from both branches, namely
$\mathcal{L}_{total} = \mathcal{L}_{rgb} + \mathcal{L}_{tir}$.

{
\setlength{\tabcolsep}{2pt}
\begin{table}[!t]
\caption{Comparison on FLIR and M3FD benchmarks. The Mod. means the modality. The best method is marked in bold, and the second-best one is marked with an underline.
}
\label{tab:cmp_flir}
\centering
\scriptsize
\begin{tabularx}{\linewidth}{
l
l
>{\centering\arraybackslash}X
c
>{\centering\arraybackslash}X
>{\centering\arraybackslash}X
c
}
\toprule
Datasets    & Method  & Mod.  & Backbone  & \makecell{Params. \\ (M)}   & \makecell{mAP \\ (\%)}  & \makecell{mAP50 \\ (\%)}    \\ \midrule
FLIR    & DINO~\cite{dino}    & RGB   & SwinT-tiny  & 48   & 32.6 & 68.6 \\
FLIR    & DINO~\cite{dino}    & TIR   & SwinT-tiny  & 48   & 41.8 & 78.7 \\ \cmidrule(lr){1-7}
FLIR    & GAFF~\cite{gaff}    & R+T   & ResNet-18 & -- & 37.5  & 72.9  \\
FLIR    & CFT~\cite{qingyun2021cft}   & R+T   & CSP-Darknet & 206  & 40.2  & 78.7  \\
FLIR    & CMX~\cite{zhang2023cmx} & R+T   & SwinT-tiny    & 181.1    & 42.3  & 82.2  \\
FLIR    & IGT~\cite{chen2023igt}  & R+T   & SwinT-tiny    & --  & \underline{43.6}  & \textbf{85}    \\
FLIR    & ICAFusion~\cite{shen2024icafusion}  & R+T   & CSP-Darknet & 120.2  & 41.4  & 79.2  \\
FLIR    & CrossFormer~\cite{lee2024crossformer} & R+T   & ResNet-50    & --   & 42.1  & 79.3  \\
FLIR    & MMFN~\cite{yang2024multidimensional}    & R+T   & ResNet-50 & 176  & 41.7  & 80.8  \\
FLIR    & RDMI~\cite{Tian_Yang_Zhu_Wang_He_2025} & R+T    & CSP-Darknet & 72.2  & 41.2     & 78.8 \\
FLIR    & MDQF~(Ours) & R+T   & SwinT-tiny  & 96  & \textbf{43.8}  & \underline{83.1}  \\ \midrule
M3FD    & DINO~\cite{dino}    & RGB   & SwinT-tiny    & 48   & 53.9  & 86.7  \\
M3FD    & DINO~\cite{dino}    & TIR   & SwinT-tiny    & 48   & 53.9  & 86.7  \\ \cmidrule(lr){1-7}
M3FD    & EME~\cite{Zhang_TIV}    & R+T   & ResNet-50 & 36.4 & 54    & 82.9  \\ 
M3FD    & FusionMamba~\cite{dong2024fusion}   & R+T   & CSP-Darknet & 287  & \textbf{61.9}  & \underline{88}    \\
M3FD    & MMFN~\cite{yang2024multidimensional}    & R+T   & ResNet-50 & 176  & -- & 86.2  \\
M3FD    & RDMI~\cite{Tian_Yang_Zhu_Wang_He_2025}  & R+T   & CSP-Darknet & 72.2 & 49.7  & 79.2  \\
M3FD    & MDQF~(Ours) & R+T   & SwinT-tiny    & 96     & \underline{55.9}  & \textbf{90.4}  \\
\bottomrule
\end{tabularx}
\end{table}
}

{
\setlength{\tabcolsep}{2pt}
\begin{table}[t]
\caption{Ablation study on FLIR benchmark. DINO is our single-branch baseline. $k_1$ and $k_2$ are \textit{k} in training and testing.}
\label{tab:ablation}
\centering
\scriptsize
\begin{tabularx}{\linewidth}{
c
>{\centering\arraybackslash}X
>{\centering\arraybackslash}X
cc
c
>{\centering\arraybackslash}X
>{\centering\arraybackslash}X
}
\toprule
Modality  & $k_1$   & $k_2$    & \makecell{Stage-specific \\ Adapter} & \makecell{Joint \\ Training} & \makecell{Post- \\ proc} & \makecell{mAP \\(\%)}  & \makecell{mAP50 \\ (\%)} \\ \midrule 
DINO-R      & 900   & 900   & --     & --     & --     & 32.6  & 68.6 \\ 
DINO-T      & 900   & 900   & --     & --     & --     & 41.8  & 78.7 \\
\cmidrule(lr){1-8}
R+T   & 900   & 900   & \ding{56} & \ding{56} & NMS   & 41.7  & 80.2 \\ 
R+T   & 900   & 900   & \ding{52} & \ding{56} & NMS   & 42.7  & 81 \\
R+T   & 900   & 900   & \ding{52} & \ding{52} & NMS   & 43.6  & 82.8 \\
R+T   & 1800  & 1800  & \ding{52} & \ding{52} & NMS   & \textbf{43.8}  & \textbf{83.1} \\
R+T   & 1800  & 1800  & \ding{52} & \ding{52} & Top-300   & 31.1  & 55.1 \\
R+T   & 1800  & 900     & \ding{52} & \ding{52} & NMS   & 43.3  & 82.1  \\
R+T   & 900   & 1800    & \ding{52} & \ding{52} & NMS   & 43.8  & 82.9  \\
R+T   & 600   & 900     & \ding{52} & \ding{52} & NMS   & 43.8  & 82.8  \\
R+T   & 300   & 450     & \ding{52} & \ding{52} & NMS   & 43.7  & 82.8  \\
\bottomrule
\end{tabularx}
\end{table}
}

\vspace{-1mm}
\section{Experiments}
\vspace{-1mm}
\subsection{Setup} \vspace{-1mm}
\textbf{Datasets.} 
The FLIR ADAS dataset is a widely used RGB-T benchmark for object detection. We conventionally use the well-registered and relabeled version provided by Zhang~\cite{zhang2020multispectral_flir} as in recent works. All the RGB-T image pairs have a resolution of 640$\times$512.
The M$^3$FD dataset~(M3FD, \cite{m3fd}) is another benchmark widely used in RGB-T image fusion and object detection, with a resolution of 1024$\times$768. We follow the conventional testing paradigm~\cite{Zhang_TIV,dong2024fusion} for a fair comparison, where 80\% of the dataset is randomly selected for training, and the remaining 20\% is reserved for testing.

\noindent
\textbf{Evaluation Settings.}
We adopt the MS-COCO style evaluation protocol throughout all experiments, using mean Average Precision~(mAP) as the performance metric. The IoU thresholds are set to 50\%~(m/AP50), and an average of multiple IoU thresholds ranging from 0.50 to 0.95~(mAP 0.5:0.95, termed mAP in our work).

\noindent
\textbf{Implementation Details}
The DETR-like detector in each single-modality branch is DINO~\cite{dino} equipped with a ViT-tiny as the backbone. Each DINO is trained independently for 12 epochs using single-modality images in the first stage. Then, in the second-stage training, the whole model is trained respectively with RGB-T image pairs for 12 and 2 epochs for the FLIR and M3FD datasets. The model is optimized with the AdamW optimizer with $lr=1e-4$. The batch size is set to 2 in all experiments.

\vspace{-1mm}
\subsection{Comparison with State-of-the-art Methods} \vspace{-1mm}

\textbf{FLIR.}
We compare our method with other state-of-the-art methods on the FLIR benchmark, and it outperforms methods with similar parameter complexity in most cases, as shown in Table~\ref{tab:cmp_flir}. Some methods may appear lightweight in terms of backbone size, but they have complex interaction modules for feature fusion, \eg~cross-attention between queries from different modalities in IGT~\cite{chen2023igt}. These mechanisms incur high computational costs and lack modality decoupling, making them susceptible to noise under degraded or imbalanced conditions. However, our approach introduces minimal additional structure and has higher modality independence. 

\noindent
\textbf{M3FD.}
The comparisons with other methods on the M3FD benchmark are also shown in Table~\ref{tab:cmp_flir}. This dataset has higher imaging quality, allowing a comparison with the image fusion–based method. EME~\cite{Zhang_TIV} is a recent RGB-T image fusion algorithm. To enable RGB-T detection, EME requires a preceding pixel-level fusion network, whose parameters are not included in the reported count. The experimental results suggest that image fusion–based methods are not an effective modality fusion strategy for the object detection task, and show the superiority of our proposed MDQF.

{
\setlength{\tabcolsep}{2pt}
\begin{table}[!t]
\caption{Modality-decoupled optimization experiments. The icons \includegraphics[height=0.45em]{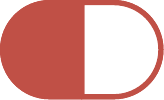}/\includegraphics[height=0.45em]{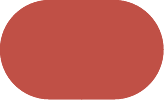} and \includegraphics[height=0.45em]{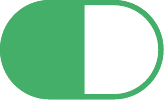}/\includegraphics[height=0.45em]{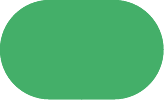} mean the 50\%/100\% RGB and 50\%/100\% TIR images of a benchmark.}
\label{tab:dcp_update}
\centering
\scriptsize
\begin{tabularx}{\linewidth}{
l
l
>{\centering\arraybackslash}X
>{\centering\arraybackslash}X
>{\centering\arraybackslash}X
}
\toprule
Datasets    & Model states & \makecell{Training data \\ setting}    & mAP~(\%)    & mAP50~(\%) \\ \midrule
FLIR & Baseline (DINO)  & \includegraphics[height=0.55em]{rgb_50.pdf} & 28.9       & 63.1       \\
FLIR & Baseline (DINO)  & \includegraphics[height=0.55em]{tir_50.pdf} & 39.7       & 75.4       \\
FLIR & MDQF (Ours)      & \includegraphics[height=0.55em]{rgb_50.pdf} + \includegraphics[height=0.55em]{tir_50.pdf} & 42.3       & 81.7       \\ \cmidrule(lr){1-5}
FLIR & DINO-high (DHR)  & \includegraphics[height=0.55em]{rgb_100.pdf} & 33.2       & 70.1       \\
FLIR & DINO-high (DHT)  & \includegraphics[height=0.55em]{tir_100.pdf} & 42.7       & 80         \\
FLIR & MDQF- w/ DHR     & \includegraphics[height=0.55em]{rgb_50.pdf} + \includegraphics[height=0.55em]{tir_50.pdf} & 42.7 (\green{+0.4}) & 81.6 (\red{-0.1}) \\
FLIR & MDQF- w/ DHT     & \includegraphics[height=0.55em]{rgb_50.pdf} + \includegraphics[height=0.55em]{tir_50.pdf} & 43.3 (\green{+1})   & 82.3 (\green{+0.6}) \\
FLIR & MDQF- w/ DHR+DHT & \includegraphics[height=0.55em]{rgb_50.pdf} + \includegraphics[height=0.55em]{tir_50.pdf} & 43.8 (\green{+1.5}) & 82.9 (\green{+1.2}) \\ \midrule
M3FD & Baseline (DINO)  & \includegraphics[height=0.55em]{rgb_50.pdf} & 46.6       & 78.3       \\
M3FD & Baseline (DINO)  & \includegraphics[height=0.55em]{tir_50.pdf} & 45.2       & 76.8       \\
M3FD & MDQF (Ours)      & \includegraphics[height=0.55em]{rgb_50.pdf} + \includegraphics[height=0.55em]{tir_50.pdf} & 48.9       & 83.6       \\ \cmidrule(lr){1-5}
M3FD & DINO-high (DHR)  & \includegraphics[height=0.55em]{rgb_100.pdf} & 53.9       & 86.7       \\
M3FD & DINO-high (DHT)  & \includegraphics[height=0.55em]{tir_100.pdf} & 53.9       & 86.7       \\
M3FD & MDQF- w/ DHR     & \includegraphics[height=0.55em]{rgb_50.pdf} + \includegraphics[height=0.55em]{tir_50.pdf} & 53.4 (\green{+4.5}) & 87.8 (\green{+4.2}) \\
M3FD & MDQF- w/ DHT     & \includegraphics[height=0.55em]{rgb_50.pdf} + \includegraphics[height=0.55em]{tir_50.pdf} & 54.1 (\green{+5.2}) & 87.9 (\green{+4.3}) \\
M3FD & MDQF- w/ DHR+DHT & \includegraphics[height=0.55em]{rgb_50.pdf} + \includegraphics[height=0.55em]{tir_50.pdf} & 55.2 (\green{+6.3}) & 88.8 (\green{+5.2}) \\
\bottomrule
\end{tabularx}
\end{table}
}

{
\setlength{\tabcolsep}{2pt}
\begin{table}[!t]
\caption{Experiments of modality decoupling in the testing. The best method is marked in bold, and the second-best one is marked with an underline.}
\label{tab:contrast_robust}
\centering
\scriptsize
\begin{tabularx}{\linewidth}{
l
l
>{\raggedright\arraybackslash}X
>{\raggedright\arraybackslash}X
>{\raggedright\arraybackslash}X
}
\toprule
\multirow{2}[2]{*}{Datasets} & \multirow{2}[2]{*}{Methods} & \multicolumn{3}{l}{mAP50~(\%)} \\ \cmidrule(r){3-5}
 & & RGB+TIR  & RGB-only  & TIR-only   \\ \midrule
FLIR & DINO~\cite{dino} & --    & \underline{68.6} & --       \\
FLIR & DINO~\cite{dino} & --    & --    & \underline{78.7}    \\
FLIR & DINO-Image       & 76.9 & 40.0 (-48.0\%)       & 65.8 (-14.4\%)  \\
FLIR & DINO-Feature     & 79.2 & 17.8 (-77.5\%)     & 71.1 (-10.2\%)  \\
FLIR & DINO-Box         & \underline{79.2} & \textbf{69.2} (\textbf{-12.6\%}) & \textbf{79.2} (\textbf{-0.0\%})  \\
FLIR & YOLOX-RGBT~\cite{Tian_Yang_Zhu_Wang_He_2025}   & 75.5 & 40.1 (-46.9\%) & 43.3 (-42.6\%)  \\
FLIR & RDMI~\cite{Tian_Yang_Zhu_Wang_He_2025}         & 78.8 & 52.5 (-33.4\%) & 75.2 (\underline{-4.6\%}) \\
FLIR & MDQF~(Ours)   & \textbf{83.1} & 68.0 (\underline{-18.2\%})  & 78.6 (-5.4\%)  \\ \midrule
M3FD & DINO~\cite{dino} & --    & \underline{86.7} & --       \\
M3FD & DINO~\cite{dino} & --    & --    & \textbf{86.7}       \\
M3FD & DINO-Image       & 85.7 & 59.8 (-30.2\%) & 37.1 (-56.7\%)  \\
M3FD & DINO-Feature     & 87.6 & 64.9 (-25.9\%) & 46.2 (-47.3\%)  \\
M3FD & DINO-Box         & \underline{88.4} & \textbf{87.4} (\textbf{-1.1\%}) & \underline{84.3} (\textbf{-4.64\%})  \\
M3FD & YOLOX-RGBT~\cite{Tian_Yang_Zhu_Wang_He_2025}   & 83.9 & 47.2 (-43.5\%) & 50.7 (-39.4\%)   \\
M3FD & RDMI~\cite{Tian_Yang_Zhu_Wang_He_2025}         & 79.2 & 38.2 (-51.8\%) & 62.1 (-21.6\%)   \\
M3FD & MDQF~(Ours)      & \textbf{90.4} & 84.9 (\underline{-6.0\%}) & 83.4 (\underline{-7.7\%})   \\
\bottomrule
\end{tabularx}
\end{table}
}

\vspace{-1mm}
\subsection{Ablation Study} \vspace{-1mm}
We conduct the ablation study for MDQF, as shown in Table~\ref{tab:ablation}, where the first two rows are single-modality baselines. The results indicate the design rules of our framework. The stage-specific adapters are better than the shared adapter across stages in feature fitting~(4th row). The second-stage joint training and NMS post-process are necessary~(5-7th rows). The remaining rows~(8-11th rows) of Table~\ref{tab:ablation} present the discussion on the selection of $k$ in Eq.~\eqref{eq:topk}. 
We find that the performance is associated with the relative magnitudes of $k_1$ and $k_2$, rather than determined by the absolute value of $k$. This characteristic helps decrease the computational cost and maintain the precision.

{
\setlength{\textfloatsep}{10pt}
\begin{figure}[!t]
    \centering
    \includegraphics[width=\linewidth]{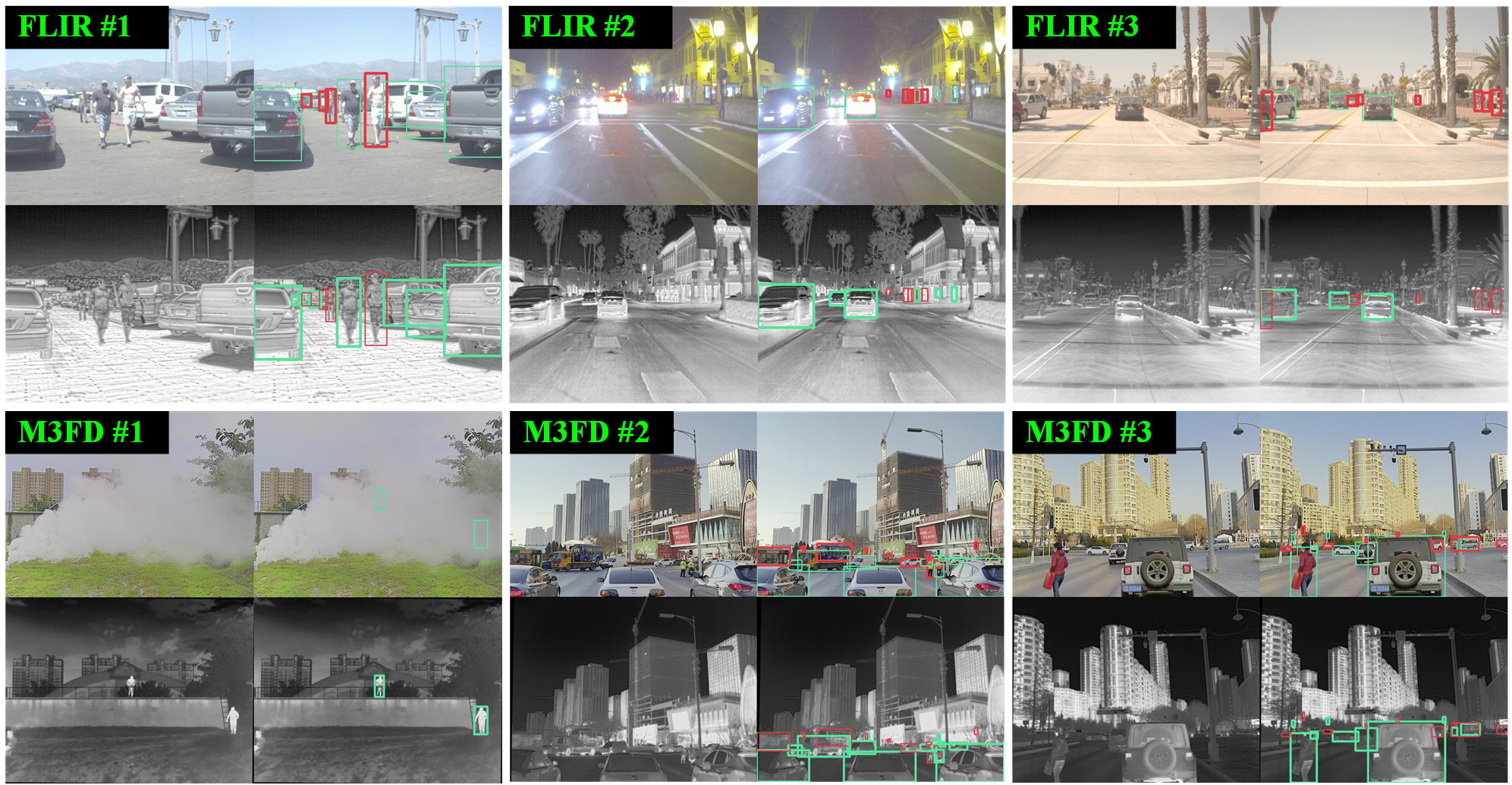}
    \captionsetup{skip=3pt}
    \caption{Qualitative results. Objects detected by the RGB branch are marked with bold red boxes in RGB images, and those from the TIR branch are marked with bold green boxes in TIR images.}
    \label{fig:qualitative}
\end{figure}
}

\vspace{-1mm}
\subsection{Analysis of Modality-decoupling} \vspace{-1mm}

\textbf{Optimization.} 
A key advantage of our proposed detection framework is that the performance can be improved by independently optimizing a single-modality branch using unpaired RGB or TIR data. Further gains are achieved through integration and joint fine-tuning with a small amount of paired RGB-T data. This is validated by the results in Table~\ref{tab:dcp_update}. In this experiment, following the proposed training paradigm, we first obtain the initial MDQF using 50\% of the paired RGB-T data from the dataset. Next, the entire~(100\%) RGB and TIR images are respectively and separately used to train two new DETR models~(denoted as DINO-high), simulating the availability of additional single-modality~(unpaired) training data. Finally, the DINO-high models are loaded into the initial MDQF, followed by joint fine-tuning using the original paired data from the first step.

\noindent
\textbf{Testing.}
Beyond enabling modality decoupling during training, our method retains this capability in prediction, thus enhancing the detector’s robustness under modality degradation during testing. We decrease the contrast of one modality to \textit{Zero} to simulate the imbalanced modality degradation. Based on DINO, we implement three traditional image-, feature-, and box-level fusion for comparison, as shown in Table~\ref{tab:contrast_robust}. Results show that our method achieves a favorable trade-off between peak performance and robustness under modality degradation. The qualitative results for two benchmarks in Figure~\ref{fig:qualitative} show the robustness of MDQF in challenging conditions.

\vspace{-1mm}
\section{Conclusion} \vspace{-1mm}

In this work, we introduce an RGB-T detector, MDQF, equipped with query fusion. Query fusion identifies and retains high-quality queries from both RGB and TIR DETR branches, excluding low-quality queries originating from degraded modalities. This mechanism enhances predictions through complementary modal prompts and prevents failures under modality degradation. 
We introduce a separate-to-joint training to reinforce the decoupling between RGB and TIR branches. MDQF attains state-of-the-art results on multiple benchmarks, preserving branch independence and enabling independent optimization with single-modality data. Under degraded image pairs, MDQF demonstrates its robustness in testing.

\clearpage
\bibliographystyle{IEEEbib}
\bibliography{refs}

@String(ICIP = {IEEE Int. Conf. Image Process.})

@String(ICIP  = {ICIP})

@inproceedings{ku2018joint,
  title={Joint 3d proposal generation and object detection from view aggregation},
  author={Ku, Jason and Mozifian, Melissa and Lee, Jungwook and Harakeh, Ali and Waslander, Steven L},
  booktitle={2018 IEEE/RSJ International Conference on Intelligent Robots and Systems (IROS)},
  pages={1--8},
  year={2018},
  organization={IEEE}
}

@article{liu2016multispectral,
  title={Multispectral deep neural networks for pedestrian detection},
  author={Liu, Jingjing and Zhang, Shaoting and Wang, Shu and Metaxas, Dimitris N},
  journal={arXiv preprint arXiv:1611.02644},
  year={2016}
}

@inproceedings{zhang2019weakly,
  title={Weakly aligned cross-modal learning for multispectral pedestrian detection},
  author={Zhang, Lu and Zhu, Xiangyu and Chen, Xiangyu and Yang, Xu and Lei, Zhen and Liu, Zhiyong},
  booktitle={Proceedings of the IEEE/CVF international conference on computer vision},
  pages={5127--5137},
  year={2019}
}

@article{kim2021uncertainty,
  title={Uncertainty-guided cross-modal learning for robust multispectral pedestrian detection},
  author={Kim, Jung Uk and Park, Sungjune and Ro, Yong Man},
  journal={IEEE Transactions on Circuits and Systems for Video Technology},
  volume={32},
  number={3},
  pages={1510--1523},
  year={2021},
  publisher={IEEE}
}

@inproceedings{gaff,
  title={Guided attentive feature fusion for multispectral pedestrian detection},
  author={Zhang, Heng and Fromont, Elisa and Lef{\`e}vre, S{\'e}bastien and Avignon, Bruno},
  booktitle={Proceedings of the IEEE/CVF winter conference on applications of computer vision},
  pages={72--80},
  year={2021}
}

@inproceedings{chen2022multimodal,
  title={Multimodal object detection via probabilistic ensembling},
  author={Chen, Yi-Ting and Shi, Jinghao and Ye, Zelin and Mertz, Christoph and Ramanan, Deva and Kong, Shu},
  booktitle={European Conference on Computer Vision},
  pages={139--158},
  year={2022},
  organization={Springer}
}

@inproceedings{mbnet,
  title={Improving multispectral pedestrian detection by addressing modality imbalance problems},
  author={Zhou, Kailai and Chen, Linsen and Cao, Xun},
  booktitle={Computer Vision--ECCV 2020: 16th European Conference, Glasgow, UK, August 23--28, 2020, Proceedings, Part XVIII 16},
  pages={787--803},
  year={2020},
  organization={Springer}
}

@inproceedings{zhang2020multispectral_flir,
  title={Multispectral fusion for object detection with cyclic fuse-and-refine blocks},
  author={Zhang, Heng and Fromont, Elisa and Lefevre, S{\'e}bastien and Avignon, Bruno},
  booktitle={2020 IEEE International conference on image processing (ICIP)},
  pages={276--280},
  year={2020},
  organization={IEEE}
}

@article{tian2023cross,
  title={Cross-Modality Proposal-guided Feature Mining for Unregistered RGB-Thermal Pedestrian Detection},
  author={Tian, Chao and Zhou, Zikun and Huang, Yuqing and Li, Gaojun and He, Zhenyu},
  journal={IEEE Transactions on Multimedia},
  year={2024},
  publisher={IEEE}
}

@article{shen2024icafusion,
  title={ICAFusion: Iterative cross-attention guided feature fusion for multispectral object detection},
  author={Shen, Jifeng and Chen, Yifei and Liu, Yue and Zuo, Xin and Fan, Heng and Yang, Wankou},
  journal={Pattern Recognition},
  volume={145},
  pages={109913},
  year={2024},
  publisher={Elsevier}
}

@article{yang2024multidimensional,
  title={Multidimensional Fusion Network for Multispectral Object Detection},
  author={Yang, Fan and Liang, Binbin and Li, Wei and Zhang, Jianwei},
  journal={IEEE Transactions on Circuits and Systems for Video Technology},
  year={2024},
  publisher={IEEE}
}

@article{lee2024crossformer,
  title={CrossFormer: Cross-guided attention for multi-modal object detection},
  author={Lee, Seungik and Park, Jaehyeong and Park, Jinsun},
  journal={Pattern Recognition Letters},
  volume={179},
  pages={144--150},
  year={2024},
  publisher={Elsevier}
}

@article{qingyun2021cft,
  title={Cross-modality fusion transformer for multispectral object detection},
  author={Qingyun, Fang and Dapeng, Han and Zhaokui, Wang},
  journal={arXiv preprint arXiv:2111.00273},
  year={2021}
}

@article{zhang2023cmx,
  title={CMX: Cross-modal fusion for RGB-X semantic segmentation with transformers},
  author={Zhang, Jiaming and Liu, Huayao and Yang, Kailun and Hu, Xinxin and Liu, Ruiping and Stiefelhagen, Rainer},
  journal={IEEE Transactions on intelligent transportation systems},
  volume={24},
  number={12},
  pages={14679--14694},
  year={2023},
  publisher={IEEE}
}

@article{dong2024fusion,
  title={Fusion-mamba for cross-modality object detection},
  author={Dong, Wenhao and Zhu, Haodong and Lin, Shaohui and Luo, Xiaoyan and Shen, Yunhang and Liu, Xuhui and Zhang, Juan and Guo, Guodong and Zhang, Baochang},
  journal={arXiv preprint arXiv:2404.09146},
  year={2024}
}

@article{chen2023igt,
  title={IGT: Illumination-guided RGB-T object detection with transformers},
  author={Chen, Keyu and Liu, Jinqiang and Zhang, Han},
  journal={Knowledge-Based Systems},
  volume={268},
  pages={110423},
  year={2023},
  publisher={Elsevier}
}

@article{yuan2024improving,
  title={Improving RGB-infrared object detection with cascade alignment-guided transformer},
  author={Yuan, Maoxun and Shi, Xiaorong and Wang, Nan and Wang, Yinyan and Wei, Xingxing},
  journal={Information Fusion},
  volume={105},
  pages={102246},
  year={2024},
  publisher={Elsevier}
}

@ARTICLE{Zhang_TIV,
  author={Zhang, Xue and Cao, Si-Yuan and Wang, Fang and Zhang, Runmin and Wu, Zhe and Zhang, Xiaohan and Bai, Xiaokai and Shen, Hui-Liang},
  journal={IEEE Transactions on Intelligent Vehicles}, 
  title={Rethinking Early-Fusion Strategies for Improved Multispectral Object Detection}, 
  year={2024},
  volume={},
  number={},
  pages={1-15},
  doi={10.1109/TIV.2024.3462488}}

@inproceedings{m3fd,
  title={Target-aware dual adversarial learning and a multi-scenario multi-modality benchmark to fuse infrared and visible for object detection},
  author={Liu, Jinyuan and Fan, Xin and Huang, Zhanbo and Wu, Guanyao and Liu, Risheng and Zhong, Wei and Luo, Zhongxuan},
  booktitle={Proceedings of the IEEE/CVF conference on computer vision and pattern recognition},
  pages={5802--5811},
  year={2022}
}

@inproceedings{detr,
  title={End-to-end object detection with transformers},
  author={Carion, Nicolas and Massa, Francisco and Synnaeve, Gabriel and Usunier, Nicolas and Kirillov, Alexander and Zagoruyko, Sergey},
  booktitle={European conference on computer vision},
  pages={213--229},
  year={2020},
  organization={Springer}
}

@article{dino,
  title={Dino: Detr with improved denoising anchor boxes for end-to-end object detection},
  author={Zhang, Hao and Li, Feng and Liu, Shilong and Zhang, Lei and Su, Hang and Zhu, Jun and Ni, Lionel M and Shum, Heung-Yeung},
  journal={arXiv preprint arXiv:2203.03605},
  year={2022}
}

@article{Tian_Yang_Zhu_Wang_He_2025, title={Learning a robust RGB-Thermal detector for extreme modality imbalance}, volume={196}, ISSN={0167-8655}, DOI={https://doi.org/10.1016/j.patrec.2025.05.005},  journal={Pattern Recognition Letters}, author={Tian, Chao and Yang, Chao and Zhu, Guoqing and Wang, Qiang and He, Zhenyu}, year={2025}, pages={1–8} }

@article{xiong2025efficient,
  title={Efficient Multispectral Object Detection with attentive feature aggregation leveraging zero-shot implicit illumination guidance},
  author={Xiong, Zhongxia and Yao, Ziying and Liu, Xuan and Zhao, Wenyao and Cao, Jie and Wu, Xinkai},
  journal={Information Fusion},
  volume={118},
  pages={102939},
  year={2025},
  publisher={Elsevier}
}

\end{document}